\documentclass{article}
\usepackage{arxiv}
\usepackage[T1]{fontenc}
\usepackage[utf8]{inputenc} 
\usepackage{booktabs}       
\usepackage{amsfonts}       
\usepackage{nicefrac}       
\usepackage{microtype}      
\usepackage{lipsum}

\usepackage{multirow}
\usepackage{graphicx}
\usepackage{amsmath}
\usepackage{hyperref}
\usepackage{color}

\newtheorem{proposition}{Proposition}    

\title{Local Multi-Label Explanations for Random Forest}

\author{
  Nikolaos Mylonas\\
    Aristotle University of Thessaloniki,\\ 54636, Greece\\
    \texttt{myloniko@csd.auth.gr}\\
    \And  
  Ioannis Mollas\\
    Aristotle University of Thessaloniki,\\ 54636, Greece\\
    \texttt{iamollas@csd.auth.gr}\\
    \And
  Nick Bassiliades\\
    Aristotle University of Thessaloniki,\\ 54636, Greece\\
    \texttt{nbassili@csd.auth.gr}\\
    \And
  Grigorios Tsoumakas\\
    Aristotle University of Thessaloniki,\\ 54636, Greece\\
    \texttt{greg@csd.auth.gr}\\
}

\begin{document}
\maketitle      
\begin{abstract}

Multi-label classification is a challenging task, particularly in domains where the number of labels to be predicted is large. Deep neural networks are often effective at multi-label classification of images and textual data. When dealing with tabular data, however, conventional machine learning algorithms, such as tree ensembles, appear to outperform competition. Random forest, being a popular ensemble algorithm, has found use in a wide range of real-world problems. Such problems include fraud detection in the financial domain, crime hotspot detection in the legal sector, and in the biomedical field, disease probability prediction when patient records are accessible. Since they have an impact on people's lives, these domains usually require decision-making systems to be explainable. Random Forest falls short on this property, especially when a large number of tree predictors are used. This issue was addressed in a recent research named LionForests, regarding single label classification and regression. In this work, we adapt this technique to multi-label classification problems, by employing three different strategies regarding the labels that the explanation covers. Finally, we provide a set of qualitative and quantitative experiments to assess the efficacy of this approach.

\end{abstract}

\section{Introduction}
Multi-label classification is a popular machine learning task, concerned with assigning multiple different labels to a single sample~\cite{tsoumakasK011}. There are plenty of applications employing multi-label classification, such as semantic indexing~\cite{PapanikolaouTLM17} and object detection~\cite{GONG2019174}. Multi-label classification has also proven useful in the predictive maintenance~\cite{aipm} and financial services~\cite{BOGAERT2019620} sectors, where tabular data are mainly used. When this sort of data is available, ensemble methods are typically outperforming other families of methods~\cite{7492171,ROKACH20147507}. Ensembles, however, are intrinsically not explainable. This is an important weakness, as explainability is useful for the vast majority of ML applications, and a necessity when they impact human lives or incur economic costs~\cite{ric,8466590}.

This paper focuses on the explainability of random forest (RF)~\cite{randomForests} models in the context of multi-label classification. There exists a lot of work on the explainability of RF for regression and single label classification tasks \cite{InTrees,defragtrees,chirps,anotsogoodpaper}. However, adapting these methods to multi-label tasks, where RF models find frequent use~\cite{Support2,SupportS,Support1}, is not straightforward. There are also techniques that have been specifically designed for multi-label tasks \cite{marlen,Tabia}. These are, however, independent of the explained model's architecture, and therefore cannot exploit the specific properties of RF models to their benefit.



To address the lack of RF-specific explainability techniques for multi-label classification in the literature, we propose an extension of LionForests (LF)~\cite{anotsogoodpaper} towards explaining multi-label classification decisions. We introduce three different strategies concerning the scope of the provided explanation (single label, predicted labelset, label subsets). We compare these strategies against similar state-of-the-art techniques, through a set of quantitative and qualitative experiments. The results highlight the conciseness of the explanations of the proposed approach.

The rest of this paper is organized as follows. Section \ref{related_work} discusses relevant research, while Section \ref{labelebackground} introduces important concepts of the LF method. Section \ref{lf_multi} presents the three novel strategies for explaining multi-label RFs. The experimental procedure, along with the data sets used, and the results, are mentioned in Section \ref{experiments}. Finally, we conclude and propose future steps for this research in Section \ref{conclusion}.

\section{Related work}
\label{related_work}

Explainability techniques can be classified into two categories, depending on their applicability to different types of models. {\em Model-agnostic} techniques ignore model structure and are therefore applicable to any ML model, whereas {\em model-specific} techniques are designed to interpret a certain family of models. The latter can either alter the model's structure to achieve explainability, or simply leverage information from the architecture without affecting it. Another distinction is between {\em global} and {\em local} explainability techniques, with the former explaining the entire model and the latter focusing on particular predictions of instances. 

We first discuss model agnostic explainability methods, which could be applied to multi-label classification, with some modifications. The use of simpler surrogate models that mimic the behavior of more complex ones, while also being more interpretable, is a topic studied in the literature and can be applied in a multi-label setting as well. Surrogates are built based on the input and the produced output of the model, and can be used to provide both global and local interpretations.

LIME \cite{lime}, the most well-known explainability technique, provides local interpretations for all types of models by estimating feature importance using perturbation methods. Similarly, Anchors \cite{anchors} extracts rule-based interpretations using a slightly different perturbation approach. Another interesting technique, LORE \cite{riccardo_guidotti1}, uses a genetic algorithm to generate neighbors, which are then used to train a decision tree that produces rule-based explanations. By constructing a decision tree to approximate the performance of complicated models, single tree approximations are used to simplify the prediction pattern of complex black-box models. TREPAN \cite{trepan} approaches this task as an inductive learning problem, aiming to represent a complex model, such as a neural network, with symbolic knowledge. This is a rule learning task, according to RuleFit \cite{rulefit}, in which each rule covers a small portion of the input space. RuleFit extracts rules from each decision tree to build a sparse linear model that incorporates both the original features and the retrieved rules, using decision trees as base learners for the various input variables. The final interpretation is based on feature importance.

We continue with explainability methods designed specifically for RF.
A model-specific approximation technique for RF called DefragTrees \cite{defragtrees} formulates the simplification of tree ensembles as a model selection problem. The aim of this work is to derive the simplest model possible that has similar performance to the whole ensemble. To do so, they employ a Bayesian model selection method to optimize the simplified model. InTrees \cite{InTrees}, on the other hand, approaches the same problem by providing a framework for selecting and pruning specific rules from the entire ensemble, effectively summarizing the relevant rules into a new simpler and more interpretable learner that can be used for future predictions. 

The RF explainability techniques discussed so far concern global explanations covering the whole model. We now move to local explanation methods. \cite{moore} introduces a local explainability method, which provides rule-based explanations exploiting feature importance. CHIRPS \cite{chirps} extracts the relevant paths for an instance from each decision tree and filters them to reduce the complexity of the explanation. LionForests (LF)~\cite{anotsogoodpaper} provides explanations for the decisions of a random forest in the form of rules. A key advantage of LF is that it distills the interpretation from the knowledge already present in RF, while also providing complete explanations. This in turn means that these explanations are provided without any demerits in the model's performance or complexity. LF can be used in binary or multi-class classification and regression problems.

Finally, we discuss RF explainability methods with a focus on visualization. iForest \cite{iForest} supports the visualization of relevant paths by multi-dimensional projection. It further allows the summarization of those paths into a final one that can be used as an explanation. Another visualization tool that can provide a global overview of a random forest model in conjunction with local explanations is ExMatrix \cite{ExMatrix}. Both the global and local explanations provided by this method come in the form of a table. 

We close the related work section with a method that has been designed specifically for multi-label classification, which is applicable to RF as well. MARLENA~\cite{marlen} is a model-agnostic approach that can provide local interpretations for black-box models by creating a neighborhood of similar instances to the one to be explained and training a decision tree. This approach is applicable to any black-box model and was evaluated mainly on health applications.

\section{LionForests}
\label{labelebackground}

This section introduces the fundamental concepts of LionForests (LF). The main step behind the interpretation extraction process of the technique, is the estimation of the minimum number of paths across the different estimators of the RF model that cover the examined instance. From each tree estimator, we extract one path responsible for the instance's prediction. Then, the set of paths that positively vote for that instance is identified. Through feature and path reduction, as well as feature-range formulation upon those paths, LF builds the interpretation. The estimation of the minimum number of paths is not a straightforward task, especially since it needs to comply with LF's main property, namely \textit{conclusiveness}. This property requires the rules produced by an explainability technique to be free of misleading or erroneous elements. 

Since a multi-label classification task can be decomposed into multiple binary classification tasks, one for each label, we will further discuss how LF computes the minimum number of paths for binary classification tasks. To better understand the following statements, we first need to mention Proposition \ref{prop1} introduced in the original paper, which is pivotal in LF's procedure.

\begin{proposition}
\label{prop1}
An RF model with a set of trees ($T$), casting $|T|$ votes, always predicts class M if and only if class M has at least a $quorum$ of votes or more, where $quorum$ $= \lfloor \frac{|T|}{2}+1 \rfloor$ out of $|T|$ votes.
\end{proposition}

Proposition \ref{prop1} states that the minimum number of paths needed for the RF model to maintain its original prediction is $ \lfloor\frac{|T|}{2}+1\rfloor$. The validity of this statement is proved in the original paper. With that in mind, the minimum number of paths which cover the examined instance that LF tries to compute is actually the quorum based on Proposition \ref{prop1}. 

Given the minimum number of paths, LF reduces the paths extracted from each RF decision tree using a sequence of procedures, obtaining the reduced set of paths extracted from the reduced trees ($T'$). In this order, they are a) reduction through association rules, b) reduction through clustering, and c) reduction by random selection. Each procedure serves a different purpose. Reduction by association rules and by clustering aim to reduce features by selecting paths with similar feature sets, whereas reduction by random selection reduces paths to the quorum.

Finally, after obtaining the reduced paths, LF identifies the common features and their ranges, merging them to obtain a single range for each feature, which we call \textit{feature-range}. The lower (upper) bound of the combined range is the maximum (minimum) of the lower (upper) bounds found across all ranges. This step is called feature aggregation. When categorical features are present, LF performs OneHot encoding on them, effectively obtaining OneHot features equal to the number of possible values the initial one had. 

These new OneHot features are handled based on the following principle: if a OneHot encoded feature is present in at least one path, with a value of 1, then the categorical feature it originates from is added to the final rule with the encoded feature as its value. For example, in a categorical feature regarding \textit{Country}, if the examined instance has the value \textit{Greece}, then the OneHot feature would be \textit{Country\_Greece}$ = 1$. In this case, we would include in the final rule this statement \textit{Country}$=$\textit{Greece}. The reduced paths cannot contain any other OneHot encoded feature of \textit{Country} with a value of 1, as that would mean that the path would not cover the instance. In contrast, if it is absent from all paths, this value of the categorical feature does not affect the outcome. In that case, LF searches for the other OneHot features originating from the same categorical one that have value of 0 and adds them to the rule as values of the original feature, which can affect the outcome. Following our example, if \textit{Country\_Japan} = 0, \textit{Country\_United-States} = 0 (or any other), appears in the reduced paths, and given that \textit{Country\_Greece}$ = 1$ does not exist in any, these features are included in the final rule in the following form \textit{Country} $\neq[$\textit{Japan, United States, ...}$]$. This procedure is called categorical feature handling.

\section{LionForest Multi-Label Explainability}
\label{lf_multi}

In multi-label classification, predictions come in vectors of size $|L|$, with $L$ denoting the set of all labels. In this case, an explanation could concern: i) the set of positively predicted labels, $L_{p} \subseteq L$, ii) subsets of it, $L_{p}' \subseteq L_{p}$, or iii) each one of its individual labels, $l \in L_{p}$. We propose three corresponding strategies that allow LF to output multi-label explanations, each calculating the quorum in a different way, based on the subset of $L_{p}$ that we want our explanation to cover.

\begin{figure}[ht]
    \centering
    \includegraphics[width=\textwidth]{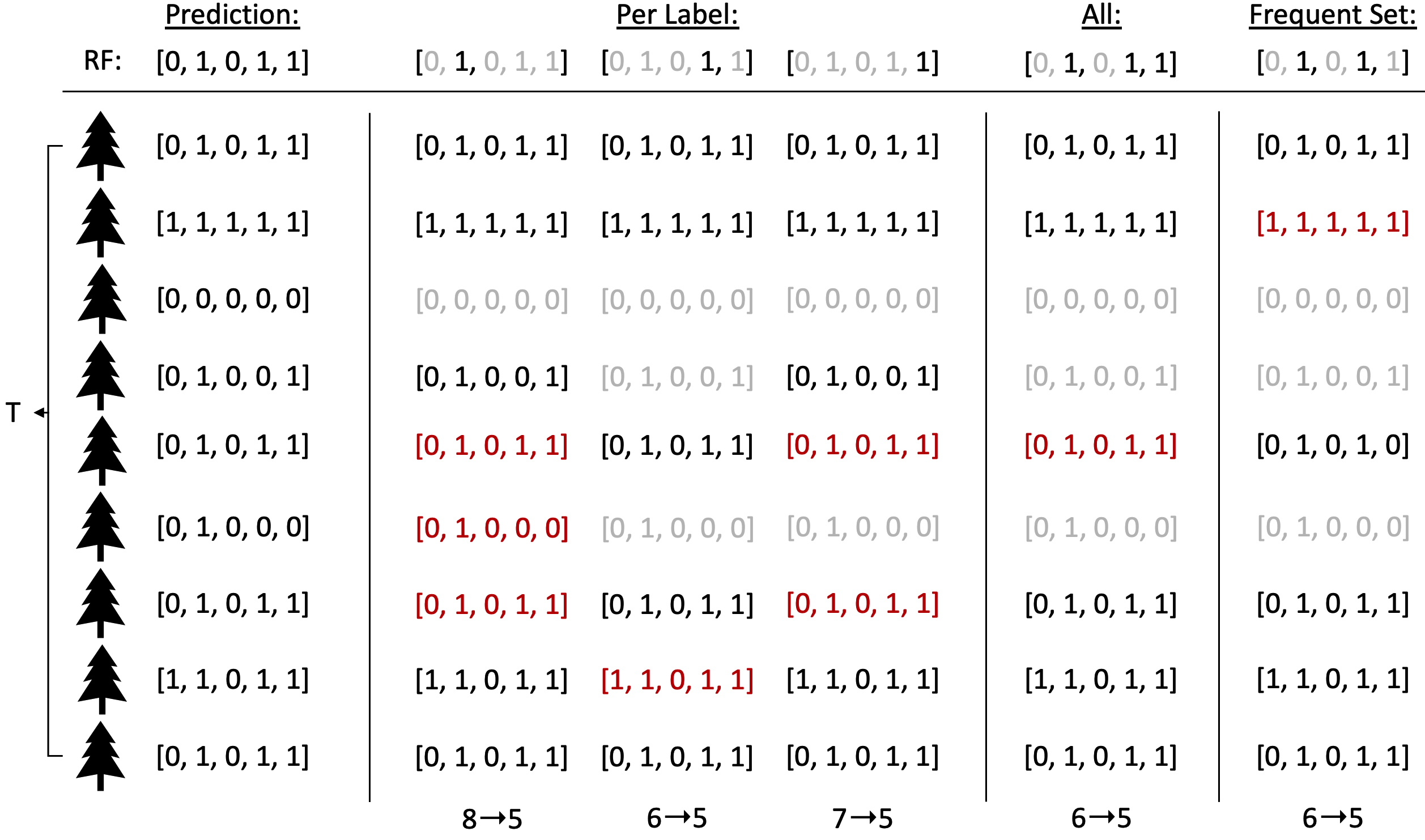}
    \caption{Running example. Greyed out predictions do not cover the examined label/labelset, whereas black and red do. The predictions whose paths were decreased are shown in red}
    \label{fig:lfmultiexample}
\end{figure}

We also introduce an example, which will be used for all strategies. Consider an RF model with $|T| = 9$ estimators, that outputs its prediction concerning $|L| = 5$ labels for a given input. Based on the theory presented before, the quorum equals $\lfloor \frac{9}{2}+1 \rfloor = \lfloor 5.5 \rfloor = 6$. For a given instance, RF predicts the following labelset $[0, 1, 0, 1, 1]$. From each $t \in T$ tree estimator, we extract the path and the prediction for this instance. Then, based on the strategy, we proceed to the appropriate reduction and eventually the formulation of the final rule interpretation. In Figure~\ref{fig:lfmultiexample}, the predicted labelsets from each $t$ tree are visible.

\subsection{Explaining each predicted label separately}

The first step of this strategy (LF-l) is the extraction of all paths regarding a decision from the tree predictors comprising the RF model. Then, for each predicted label $l \in L_p$, a multi-stage process takes place, which first identifies the paths that vote for its prediction. The next step is the reduction of $T$, as explained in Section \ref{labelebackground}, to the number denoted by the quorum, obtaining the minimum number of paths from $T'$ trees. The rule building steps remain the same as those in the original technique, namely feature aggregation and handling of categorical features. After formulating a rule for each predicted label $l$, we use these rules as an explanation for the examined instance.

In columns 2 to 4 (\textit{Per Label}) in Figure~\ref{fig:lfmultiexample}, we can see how LF selects and reduces the paths to the quorum. It identifies the paths that voted for each of the three predicted labels (black and red font). If the number of paths exceeds the quorum, the LF reduction strategies are used to decrease them to the bare minimum (black font). Treating each label separately can result in smaller feature sets in the final interpretation. This is because LF has a greater number of possible paths to reduce to the minimum.

\subsection{Explaining all the predicted labelset}

This strategy is largely similar to the previous one, with the main difference being that instead of an iterative process for each label (\textit{LF-a}), this time a single process is executed for the whole predicted labelset. This in turn means that LF must now identify the paths from the $T$ trees that include the whole predicted labelset in its prediction, greatly limiting the number of available paths to be reduced in the following step, if possible. Furthermore, during the path reduction step, each produced path set must cover the whole prediction, restricting the number of paths LF can safely remove to obtain $|T'|$. 

It is worth noting that, due to the above conditions, the final rule obtained after applying the rule-building steps is very specific to the examined instance. There is a possibility that the number of recognized paths covering all predicted labels will be less than the quorum. This prevents us from further decreasing them, but also prohibits us from using them alone to form the final rule. In this scenario, regardless of their vote, we use all the paths.

Connecting this strategy with the running example of Figure~\ref{fig:lfmultiexample}, we focus on the third column, \textit{All}. Only 6 paths include all the predicted labels at the same time. LF will use the reduction strategies to decrease those pathways to 5 (quorum). However, because there is so little room for reduction, the strategy's effectiveness is limited, and therefore, we might not observe the desired feature reduction.

\subsection{Explaining frequent label subsets}

This strategy provides explanations for subsets (\textit{LF-p}) of the predicted labelset that frequently appear inside the examined data set. These subsets are identified with the use of association rules and specifically the \textit{fpgrowth} algorithm. Then, a process comparable to the one present in the first strategy is performed. For each subset, the paths that vote for all the labels present inside it are identified and then reduced to $|T'|$, before the rule building steps that formulate the final rule for this subset are implemented. The final explanation for the frequent subsets is an aggregation of the rules built by the aforementioned process. 

In case of larger labelsets, as well as a large set of predicted labels, the number of activated subsets can be very high. Therefore, the end-user is given an option to limit the number of subsets. Hence, if the activated subsets are $X$ and the user asks for $N < X$, the first $N$ subsets and their explanation will be provided, ordered based on the support of the subset across the labelsets of the training data set.

In the example (Figure~\ref{fig:lfmultiexample}), the last column presents the explanation of one identified subset $[0,1,0,1,0] \subset [0,1,0,1,1]$, the paths which cover this set, and the removed path.

\section{Experiments}
\label{experiments}
This section summarizes the experiments that we carried out to compare the performance of our strategies to state-of-the-art techniques frequently used in the literature. To further the reliability of our results we performed a 10-fold cross validation and present the standard deviation of our showcased results in each table. We performed three distinct sets of experiments. The first compares our various strategies to each other, in order to gain insight into their effectiveness with multi-label data. The second focuses on techniques that explain the entire predicted labelset, pitting our second strategy against MARLENA, due to its relevance to our task, and two baselines: local (LS) and global (GS) tree surrogates. The third and final set compares our first strategy to Anchors and CHIRPS. Anchors was selected for its prominence in the literature, and CHIRPS for its similarity to our approach, concerning the per label experiments. Both techniques have been adjusted to provide explanations for each of the predicted labels. The code for these experiments is available in GitHub\footnote{\url{https://tinyurl.com/c5y8uxm4}} and DockerHub\footnote{\url{https://tinyurl.com/2nh4zyj3}}.

\subsection{Data sets}
Here, we present the data sets used and the pre-processing steps we followed for each one of them. We selected four multi-label tabular data sets, whose main statistics can be found in Table \ref{datasets:tab}. The values inside the parentheses concern the numbers after the pre-processing steps for each data set. 

\begin{table}[ht]
\centering
\caption{Data set information}
\label{datasets:tab}
\begin{tabular}{cccccc}
\hline
 ID & Dataset            & Instances & Features & Labels & Cardinality \\ \hline
D1 & Food Truck    & 407       & 21 (29)  & 12     & 2.29       \\
D2 & Water Quality  & 1060      & 16       & 14     & 5.07       \\

D3 & Flags         & 194       & 19       & 7      & 3.39       \\

D4 & AI4I           & 10K (339)       & 7 (6)    & 6 (4)  & 1.04       \\ \hline
\end{tabular}
\end{table}

\paragraph{D1: Food Truck.} This data set was compiled from the replies of 407 survey participants, concerning their food preferences, personal information, time of meal (input variables) and types of food trucks they eat from (target variables)~\cite{foodtruck}. We replaced missing values with the mean value across their column. Furthermore, four categorical features were present. As such, two of them were handled as ordinal (gender and time of the day) and the rest were one-hot encoded (motivation and marital status). Finally, the data were min-max scaled.

\paragraph{D2: Water Quality.} This scientific data set was used for modeling the quality of water in Slovenian rivers. It contains features like the water's temperature, pH, and concentration of different chemicals. The target variables correspond to different water quality indicators~\cite{waterquality}. The pre-processing was minimal only including min-max scaling.

\paragraph{D3: Flags.} The third data set used in our experiments contains information about certain countries and their culture. The target variables concern the colors that are present on the flag of each country~\cite{uci}. No pre-processing took place for this data set other than min-max scaling.

\paragraph{D4: AI4I.} The last data set employed in our experiments is a synthetic predictive maintenance data set concerning the prediction of certain types of machine failures~\cite{aipm}. Aiming to emulate real predictive maintenance problems encountered in the industry, the data set contains features like temperature, torque, rotational speed of different machines to predict possible failures. There is a single categorical feature (type) inside the data set which we handle as ordinal, and a feature without information (product ID). We then min-max scale all features. Regarding the labels, there is a strong dependency between the first denoting whether there is a failure or not and the rest of the labelset, which are the types of failures. As such, we only keep the examples exhibiting some kind of failure, while removing the first. Finally, we also remove the final label which denotes a random failure, of which very few instances exist in the data set.

\subsection{Quantitative experiments}

In order to perform our quantitative experiments, we made use of four different metrics, each one covering a different aspect of a produced rule. Rule length (L) denotes the number of feature-ranges present inside the rule. Smaller lengths are easier for the end-user to comprehend but can also indicate that the explanation is problematic. Furthermore, expert users tend to prefer larger rules, due to their richer information~\cite{bigrules}. It is worth noting that for techniques providing rules for each label or label subset separately, the final rule length is the sum of the lengths of those individual rules. Coverage (C) describes the average number of instances each rule satisfies. Precision (P) refers to the fraction of correctly covered instances among the ones covered by each rule. Higher values on both metrics correspond to better performance. Time response (T) describes the run time of the technique in seconds.


We first discuss the results of the comparison among our proposed strategies that can be seen in Table \ref{tab:self_comparison}. The goal of this setup was to compare the length of the rules produced by each strategy, along with the time needed to produce them. As expected, the strategy providing one rule for the whole predicted labelset results in shorter rules and smaller run times. This can be seen in all four data sets, albeit the differences are minimal in the last one. The frequent pair strategy seems to be the one with the longest run times and lengthiest rules. Such outcome was anticipated, given that in most cases the number of frequent subsets present in a labelset is higher than the number of distinct labels comprising it.

\begin{table}[ht]
\centering
\caption{Comparison between the different strategies of LF}
\label{tab:self_comparison}
\begin{tabular}{cccc}
\hline
Datasets            & Algorithm & L                     & T\\ \hline
\multirow{3}{*}{D1} & LF-a      & 20.33$_{\pm{0.70}}$   & 6.25$_{\pm{0.96}}$ \\
                    & LF-l      & 30.06$_{\pm{3.47}}$   & 10.30$_{\pm{0.86}}$ \\
                    & LF-p      & 44.71$_{\pm{9.04}}$   & 11.33$_{\pm{1.16}}$ \\ \hline
\multirow{3}{*}{D2} & LF-a      & 16.37$_{\pm{0.40}}$   & 0.94$_{\pm{0.04}}$ \\
                    & LF-l      & 57.64$_{\pm{5.71}}$   & 1.37$_{\pm{0.11}}$ \\
                    & LF-p      & 153.34$_{\pm{16.76}}$ & 2.87$_{\pm{0.29}}$ \\ \hline
\multirow{3}{*}{D3} & LF-a      & 18.24$_{\pm{0.49}}$   & 1.05$_{\pm{0.10}}$ \\
                    & LF-l      & 62.58$_{\pm{5.24}}$   & 5.66$_{\pm{0.40}}$ \\
                    & LF-p      & 111.72$_{\pm{7.32}}$  & 7.38$_{\pm{0.48}}$ \\ \hline
\multirow{3}{*}{D4} & LF-a      & 4.92$_{\pm{0.46}}$    & 1.38$_{\pm{0.30}}$ \\
                    & LF-l      & 5.04$_{\pm{0.48}}$    & 1.41$_{\pm{0.31}}$ \\
                    & LF-p      & 5.04$_{\pm{0.48}}$    & 1.41$_{\pm{0.30}}$ \\ \hline
\end{tabular}
\end{table}

The comparison regarding the techniques providing explanations for the whole predicted labelset can be seen in Table \ref{tab:my_all_comp}. Exploring each metric separately reveals that all four techniques evaluated have rather poor Coverage, with MARLENA having the best performance and LF-a having the worst. In terms of Precision, LF-a always achieves perfect precision due to its conclusiveness property, while both surrogate models surpass MARLENA. In all datasets, LF-a produces the lengthiest rules, while MARLENA produces the shortest. Finally, save for LF-a's poor performance on the first dataset, there is no clear winner among the other local techniques in terms of time response.

\begin{table}[ht]
\centering
\caption{Comparison between techniques explaining the whole predicted labelset}
\label{tab:my_all_comp}
\begin{tabular}{ccccccc}
\hline
Dataset             & Algorithms & C           & L         & P          & T       \\ \hline
\multirow{4}{*}{D1} & LF-a       & 0.02$_{\pm{0.00}}$ & 20.32$_{\pm{0.65}}$ & 1.00$_{\pm{0.00}}$ & 6.25$_{\pm{0.98}}$ \\
                    & GS         & 0.06$_{\pm{0.03}}$ &  7.81$_{\pm{1.39}}$ & 0.77$_{\pm{0.08}}$ & 0.00$_{\pm{0.00}}$ \\
                    & LS         & 0.07$_{\pm{0.02}}$ &  6.96$_{\pm{0.59}}$ & 0.75$_{\pm{0.07}}$ & 1.52$_{\pm{0.02}}$ \\
                    & MARLENA         & 0.15$_{\pm{0.04}}$ &  4.42$_{\pm{0.27}}$ & 0.73$_{\pm{0.05}}$ & 2.40$_{\pm{0.02}}$ \\ \hline
\multirow{4}{*}{D2} & LF-a       & 0.01$_{\pm{0.00}}$ & 16.36$_{\pm{0.40}}$ & 1.00$_{\pm{0.00}}$ & 0.94$_{\pm{0.04}}$ \\
                    & GS         & 0.02$_{\pm{0.00}}$ &  8.78$_{\pm{0.36}}$ & 0.66$_{\pm{0.05}}$ & 0.00$_{\pm{0.00}}$ \\
                    & LS         & 0.03$_{\pm{0.00}}$ &  6.67$_{\pm{0.20}}$ & 0.65$_{\pm{0.03}}$ & 1.41$_{\pm{0.03}}$ \\
                    & MARLENA         & 0.05$_{\pm{0.01}}$ &  5.97$_{\pm{0.33}}$ & 0.62$_{\pm{0.05}}$ & 2.04$_{\pm{0.01}}$ \\ \hline
\multirow{4}{*}{D3} & LF-a       & 0.05$_{\pm{0.00}}$ & 18.24$_{\pm{0.48}}$ & 1.00$_{\pm{0.00}}$ & 1.05$_{\pm{0.10}}$ \\
                    & GS         & 0.08$_{\pm{0.01}}$ &  5.22$_{\pm{0.65}}$ & 0.90$_{\pm{0.05}}$ & 0.00$_{\pm{0.00}}$ \\
                    & LS         & 0.08$_{\pm{0.02}}$ &  5.31$_{\pm{0.62}}$ & 0.91$_{\pm{0.03}}$ & 3.41$_{\pm{0.06}}$ \\
                    & MARLENA         & 0.13$_{\pm{0.03}}$ &  4.41$_{\pm{0.20}}$ & 0.85$_{\pm{0.05}}$ & 0.86$_{\pm{0.00}}$ \\ \hline
\multirow{4}{*}{D4} & LF-a       & 0.03$_{\pm{0.00}}$ &  4.92$_{\pm{0.46}}$ & 1.00$_{\pm{0.00}}$ & 1.38$_{\pm{0.30}}$ \\
                    & GS         & 0.24$_{\pm{0.15}}$ &  3.62$_{\pm{0.57}}$ & 0.92$_{\pm{0.07}}$ & 0.00$_{\pm{0.00}}$ \\
                    & LS         & 0.24$_{\pm{0.14}}$ &  3.41$_{\pm{0.38}}$ & 0.94$_{\pm{0.04}}$ & 1.43$_{\pm{0.04}}$ \\
                    & MARLENA         & 0.33$_{\pm{0.13}}$ &  2.52$_{\pm{0.26}}$ & 0.75$_{\pm{0.06}}$ & 0.80$_{\pm{0.01}}$ \\ \hline
\end{tabular}
\end{table}

\begin{table}[ht]
\centering
\caption{Comparison between techniques explaining each predicted label separately}
\label{tab:single-labele-comparison}
\resizebox{0.74\textwidth}{!}{
\begin{tabular}{cccccc}
\hline
Dataset             & Algorithms & C                  &  L                  & P                  & T         \\ \hline
\multirow{3}{*}{D1} & LF-l       & 0.03$_{\pm{0.00}}$ & 41.85$_{\pm{4.8}}$   & 1.00$_{\pm{0.00}}$ & 10.26$_{\pm{0.90}}$   \\
                    & Anchors         & 0.06$_{\pm{0.02}}$ & 8.95$_{\pm{2.3}}$   & 0.99$_{\pm{0.02}}$ & 341.62$_{\pm{105.93}}$ \\
                    & CHIRPS         & 0.73$_{\pm{0.09}}$ & 3.66$_{\pm{1.1}}$   & 0.89$_{\pm{0.03}}$ &   8.47$_{\pm{0.35}}$   \\ \hline
\multirow{3}{*}{D2} & LF-l       & 0.01$_{\pm{0.00}}$ & 57.64$_{\pm{5.71}}$ & 1.00$_{\pm{0.00}}$ &   1.37$_{\pm{0.11}}$   \\
                    & Anchors         & 0.04$_{\pm{0.01}}$ & 31.05$_{\pm{3.47}}$ & 0.96$_{\pm{0.02}}$ & 302.07$_{\pm{44.43}}$  \\
                    & CHIRPS         & 0.38$_{\pm{0.05}}$ & 10.36$_{\pm{1.37}}$ & 0.76$_{\pm{0.05}}$ &   8.24$_{\pm{0.50}}$   \\ \hline
\multirow{3}{*}{D3} & LF-l       & 0.05$_{\pm{0.00}}$ & 62.58$_{\pm{5.24}}$ & 1.00$_{\pm{0.00}}$ &   5.77$_{\pm{0.42}}$   \\
                    & Anchors         & 0.31$_{\pm{0.05}}$ &  7.36$_{\pm{1.24}}$ & 0.97$_{\pm{0.02}}$ &  63.00$_{\pm{17.91}}$  \\
                    & CHIRPS         & 0.71$_{\pm{0.05}}$ &  2.09$_{\pm{0.45}}$ & 0.94$_{\pm{0.03}}$ &   2.35$_{\pm{0.19}}$   \\ \hline
\multirow{3}{*}{D4} & LF-l       & 0.03$_{\pm{0.00}}$ &  5.04$_{\pm{0.48}}$ & 1.00$_{\pm{0.00}}$ &   1.41$_{\pm{0.31}}$   \\
                    & Anchors         & 0.14$_{\pm{0.05}}$ &  3.98$_{\pm{0.52}}$ & 0.92$_{\pm{0.07}}$ &  29.53$_{\pm{3.42}}$   \\
                    & CHIRPS         & 0.33$_{\pm{0.14}}$ &  2.44$_{\pm{0.66}}$ & 0.93$_{\pm{0.06}}$ &   1.25$_{\pm{0.09}}$   \\ \hline
\end{tabular}}
\end{table}
The last set of experiments concerns explanations for each predicted label and can be seen in Table \ref{tab:single-labele-comparison}. This setup allows us to use single-label interpretation techniques like Anchors and CHIRPS. In all datasets, CHIRPS outperforms its competitors in terms of Coverage. The longest rules are provided by LF-l, while the shortest are provided by CHIRPS. In terms of precision, LF-l works flawlessly, with Anchors ranking second with a small advantage over CHIRPS. Regarding the time response, LF-l and CHIRPS produce similar results with small deviations. Anchors, on the other hand, requires a significant amount of computational resources, leading to longer time responses.

\subsection{Qualitative experiments}

Our qualitative experiments focus on the AI4I (D4) data set, as its small feature set makes it easier to present and analyze an example. The features available in this data set along with their ranges and the values of a sample instance can be found in Table \ref{tab:features}.

\begin{table}[ht]
\centering
\caption{Dataset features, their ranges, and the values of a sample instance}
\label{tab:features}
\begin{tabular}{lcc}
\hline
\textbf{Feature}        & \textbf{Range}   & \textbf{Instance values} \\ \hline
Type                    & $[1, 3]$         & $3$ \\
Air temperature [K]     & $[295.6, 304.4]$ & $300.7$ \\
Process temperature [K] & $[306.1, 313.7]$ & $310.2$ \\
Rotational speed [rpm]  & $[1212, 2874]$   & $1364$ \\
Torque [Nm]             & $[4.2, 76.2]$    & $65.3$ \\
Tool wear [min]         & $[0, 251]$       & $208$ \\ \hline
\end{tabular}
\end{table}

\begin{table}[ht]
\centering
\caption{Example Rules by the proposed strategies}
\label{tab:strategies_rules}
\begin{tabular}{lc}
\hline
Techniques & Interpretation\\ \hline
LF-a       & \begin{tabular}[c]{@{}c@{}}
\textbf{If} $2.5 \leq Type\leq3$ and $300.6 \leq Air$ $temperature\leq301.7$ and \\ $310\leq Process$ $temperature \leq 313.7$ and \\ $1351 \leq Rotational$ $speed \leq 1380$ and $65.2 \leq Torque \leq 65.5$ and \\ $207.5 \leq Tool$ $wear \leq 209$ \textbf{then} TWF PWF OSF \end{tabular}\\ \hline
LF-l       & \begin{tabular}[c]{@{}c@{}}
\textbf{If} $2.5\leq Type \leq3$ and $295.6 \leq Air$ $temperature \leq 301.7$ and \\ $1322.5 \leq Rotational$ $speed \leq 1419.5$ and $65.2 \leq Torque \leq 76.2$ and \\ $206.5 \leq Tool$ $wear \leq 251.0$ \textbf{then} TWF \\
\textbf{If} $2.5 \leq Type \leq 3$ and $295.6 \leq Air$ $temperature \leq 301.7$ and \\ $1351 \leq Rotational$ $speed \leq 1380$ and $65.2 \leq Torque \leq 76.2$ and \\ $188 \leq Tool$ $wear \leq 251$ \textbf{then} PWF \\
\textbf{If} $2.5 \leq Type \leq3$ and $300.6 \leq Air$ $temperature \leq 300.8$ and \\ $65.2 \leq Torque \leq 65.5$ and $207.5 \leq Tool$ $wear \leq 251$ \textbf{then} OSF
\end{tabular} \\ \hline

LF-p       & \begin{tabular}[c]{@{}c@{}}
\textbf{If} $2.5 \leq Type \leq 3$ and $295.6 \leq Air$ $temperature\leq 301.7$ and \\ $1351 \leq Rotational$ $speed \leq 1380$ and $ 65.2 \leq Torque \leq 76.2$ and \\ $185 \leq Tool$ $wear \leq 251$ \textbf{then} PWF OSF
\end{tabular} \\ \hline
\end{tabular}
\end{table}

The first qualitative comparison found in Table \ref{tab:strategies_rules} includes the rules produced by the different strategies of LF for the examined instance. As the quantitative experiments in Table \ref{tab:self_comparison} suggested, we can see that LF-a, the strategy explaining all predicted labels, provides the lengthier and more specific individual rule. When LF-l is employed, we can see that for the predicted label `OSF', the rule is 2 feature-ranges smaller, while the rule regarding the predictions `PWF' and `TWF' are 1 feature-range smaller and have wider ranges. The third strategy, LF-p, produces rules explaining frequent label subsets. In this example, it produces a rule explaining the labels `PWF' and `OSF', a frequent labelset present in the prediction, and the rule is 1 feature-range smaller. Therefore, the user can choose between the available strategies based on their needs. We should mention that all these rules are conclusive, therefore, any change on the features between the given ranges, or any change on the features not appearing in the rules will not impact the prediction.

Continuing our comparisons, Table \ref{tab:rules_all} presents the explanations given by GS, LS, and MA for the whole predicted labelset. These techniques provide rules that are substantially shorter and have broader feature-ranges than LF-a. However, the trade-off for these properties is the loss of conclusiveness. To support our claim, we perform three separate modifications on the values of the examined instance (one feature at a time), to demonstrate that the rules provided by the competitors do not account for these changes, in contrast to LF-a. GS does not contain a feature-range for $Type$, suggesting that it does not impact the prediction. Nonetheless, changing the value of this feature to either $1$ or $2$ alters the prediction to `OSF'. The feature-range given by LS for $Torque$ is deceptively wide. Lowering the value of this feature from $65.3$ to $50$ changes the prediction like before. Additionally, the range displayed by MA regarding $Air$ $temperature$ is inaccurate, as increasing its value from $300.7$ to $302$ causes the prediction to change from `TWF', `PWF', and `OSF' to `OSF'. Finally, both LS and MA incorrectly ignore a feature, $Air$ $temperature$ and $Torque$, respectively, since both affect the prediction as seen before.

\begin{table}[ht]
\centering
\caption{Example rules for the whole labelset}
\label{tab:rules_all}
\begin{tabular}{cc}
\hline
~~~Technique~~~ & Interpretation\\ \hline
GS         & \begin{tabular}[c]{@{}c@{}}\textbf{If} $Air$ $temperature \leq 301.7$ and $Tool$ $wear > 176.5$ and \\ $Torque > 65.2$ \textbf{then} TWF PWF OSF\end{tabular}\\ \hline
LS         & \begin{tabular}[c]{@{}c@{}}\textbf{If} and $Tool$ $wear> 188$ and $Torque > 48.4$ and \\ $Type > 2.5$ \textbf{then} TWF PWF OSF\end{tabular}\\\hline
MARLENA         & \begin{tabular}[c]{@{}c@{}}\textbf{If} $Air$ $temperature \leq 303$ and
$Rotational$ $speed \leq 1382.4$ and \\ $Type > 2.97$ \textbf{then} TWF PWF OSF\end{tabular} \\ \hline       
\end{tabular}
\end{table}

\begin{table}[ht]
\centering
\caption{Example rules per label}
\label{tab:rules_per_label}
\begin{tabular}{cc}
\hline
Technique & Interpretation\\ \hline
Anchors         & \begin{tabular}[c]{@{}c@{}}

\textbf{If} $Air$ $temperature \leq 301.6$ and $Type > 2$ and
$Tool$ $wear > 207.5$ and \\ $Torque > 61.2$ and $Rotational$ $speed \leq 1365$ \textbf{then} TWF\\ 
\textbf{If} $Torque > 61.2$ and $Air$ $temperature \leq 301.6$ and $Type > 2$ and \\ $ 1326.5 < Rotational$ $speed \leq 1365$ and \\ $309.5 <Process$ $temperature \leq 311.2$ \textbf{then} PWF\\ 
\textbf{If} $Tool$ $wear > 207.5$ and $Torque > 61.2$ \textbf{then} OSF\end{tabular} \\ \hline

CHIRPS         & \begin{tabular}[c]{@{}c@{}}\textbf{If} \{\} \textbf{then} TWF\\ \textbf{If} $Air$ $temperature \leq 302.5$ and $Torque > 65$ \textbf{then} PWF \\ \textbf{If} $Tool$ $wear > 176.5$ and $Torque > 65$ \textbf{then} OSF\end{tabular}\\ \hline
\end{tabular}
\end{table}

Concluding our qualitative study, we present the rules provided by Anchors and CHIRPS in Table \ref{tab:rules_per_label}. We can see that Anchors provides rules of similar length to LF-l, compared to the significantly shorter ones of CHIRPS. Nevertheless, similarly to before, the shorter rules are inconclusive. Rules provided by CHIRPS have a lot of inaccuracies, with the most obvious being the empty rule for the `TWF' prediction. We also spot few inaccuracies in the lengthier rules of Anchors. For example, the explanations for labels `TWF' and `PWF' suggest that values above $61.2$  for $Torque$ lead to predictions containing these 2 labels. However, increasing the value to $65$ results in both not being predicted by the model. Contrarily, the rules provided by LF-l contain the correct feature ranges for $Torque$.


\section{Conclusion}
\label{conclusion}
This paper proposed three different strategies that extend LF, so it can be used to provide explanations for multi-label classification problems as well. Each of these strategies, explain the predicted labelset from a different point of view, resulting in rules of different length for each one of them, as well as different time responses. All three, however, retain the conclusiveness property of the original technique, providing concise explanations. 

This was validated by our experimental procedure, where all strategies achieve a Precision of 1 throughout all the different setups and data sets, something that no other competitor manages to reach. Having said that, our strategies tend to produce lengthier rules that cover a smaller portion of instances than the competitors, meaning they are more specific. This attribute is not necessarily a shortcoming, considering that expert users prefer longer, more informative explanations. In addition, the low rule length of the other techniques can be misleading, as they tend to provide even empty explanations as showcased in our qualitative experiments. 

Some of the future steps of this research include the extension of the technique, so it can also be applied to multi target regression problems, in conjunction to an extensive experimental procedure including new competitors, more data sets and additional metrics. Furthermore, a user study to assess the quality of the rules the three different strategies produce, can be performed. Doing so, will allow us to obtain insight from different types of users about the strategies and their applicability in various domains. Finally, applying a similar strategy to produce explanations for other ensemble models can also be explored in another work.

\section*{Acknowledgments}
The research work was supported by the Hellenic Foundation for Research and Innovation (H.F.R.I.) under the ``First Call for H.F.R.I. Research Projects to support Faculty members and Researchers and the procurement of high-cost research equipment grant'' (Project Number: 514)

\bibliographystyle{unsrt} 

\begin{thebibliography}{10}

\bibitem{tsoumakasK011}
Grigorios Tsoumakas and Ioannis Katakis.
\newblock Multi-label classification: An overview.
\newblock {\em Int. J. Data Warehous. Min.}, 3(3):1--13, 2007.

\bibitem{PapanikolaouTLM17}
Yannis Papanikolaou, Grigorios Tsoumakas, Manos Laliotis, Nikos Markantonatos,
  and Ioannis~P. Vlahavas.
\newblock Large-scale online semantic indexing of biomedical articles via an
  ensemble of multi-label classification models.
\newblock {\em J. Biomed. Semant.}, 8(1):43:1--43:13, 2017.

\bibitem{GONG2019174}
Tao Gong, Bin Liu, Qi~Chu, and Nenghai Yu.
\newblock Using multi-label classification to improve object detection.
\newblock {\em Neurocomputing}, 370:174--185, 2019.

\bibitem{aipm}
Stephan Matzka.
\newblock Explainable artificial intelligence for predictive maintenance
  applications.
\newblock In {\em 2020 Third International Conference on Artificial
  Intelligence for Industries (AI4I)}, pages 69--74. IEEE, 2020.

\bibitem{BOGAERT2019620}
Matthias Bogaert, Justine Lootens, Dirk {Van den Poel}, and Michel Ballings.
\newblock Evaluating multi-label classifiers and recommender systems in the
  financial service sector.
\newblock {\em European Journal of Operational Research}, 279(2):620--634,
  2019.

\bibitem{7492171}
Qingyao Wu, Mingkui Tan, Hengjie Song, Jian Chen, and Michael~K. Ng.
\newblock Ml-forest: A multi-label tree ensemble method for multi-label
  classification.
\newblock {\em IEEE Transactions on Knowledge and Data Engineering},
  28(10):2665--2680, 2016.

\bibitem{ROKACH20147507}
Lior Rokach, Alon Schclar, and Ehud Itach.
\newblock Ensemble methods for multi-label classification.
\newblock {\em Expert Systems with Applications}, 41(16):7507--7523, 2014.

\bibitem{ric}
Riccardo Guidotti, Anna Monreale, Salvatore Ruggieri, Franco Turini, Fosca
  Giannotti, and Dino Pedreschi.
\newblock A survey of methods for explaining black box models.
\newblock {\em ACM Comput. Surv.}, 51(5), aug 2018.

\bibitem{8466590}
Amina Adadi and Mohammed Berrada.
\newblock Peeking inside the black-box: A survey on explainable artificial
  intelligence (xai).
\newblock {\em IEEE Access}, 6:52138--52160, 2018.

\bibitem{randomForests}
Leo Breiman.
\newblock Random forests.
\newblock {\em Machine Learning}, 45(1):5--32, Oct 2001.

\bibitem{InTrees}
Houtao Deng.
\newblock Interpreting tree ensembles with intrees.
\newblock {\em Int. J. Data Sci. Anal.}, 7(4):277--287, 2019.

\bibitem{defragtrees}
Satoshi Hara and Kohei Hayashi.
\newblock Making tree ensembles interpretable: A bayesian model selection
  approach.
\newblock In Amos Storkey and Fernando Perez-Cruz, editors, {\em Proceedings of
  the Twenty-First International Conference on Artificial Intelligence and
  Statistics}, volume~84 of {\em Proceedings of Machine Learning Research},
  pages 77--85. PMLR, 09--11 Apr 2018.

\bibitem{chirps}
Julian Hatwell, Mohamed~Medhat Gaber, and R.~Muhammad~Atif Azad.
\newblock {CHIRPS}: Explaining random forest classification.
\newblock {\em Artificial Intelligence Review}, 53(8):5747--5788, June 2020.

\bibitem{anotsogoodpaper}
Ioannis Mollas, Nick Bassiliades, and Grigorios Tsoumakas.
\newblock Conclusive local interpretation rules for random forests.
\newblock {\em CoRR}, abs/2104.06040, 2021.
\newblock to appear in DAMI, Springer.

\bibitem{Support2}
Samaneh Kouchaki, Yang Yang, Alexander Lachapelle, Timothy~M. Walker, A.~Sarah
  Walker, CRyPTIC~Consortium ~, Timothy E.~A. Peto, Derrick~W. Crook, and
  David~A. Clifton.
\newblock Multi-label random forest model for tuberculosis drug resistance
  classification and mutation ranking.
\newblock {\em Frontiers in Microbiology}, 11, 2020.

\bibitem{SupportS}
Seema Sharma and Deepti Mehrotra.
\newblock Comparative analysis of multi-label classification algorithms.
\newblock In {\em 2018 First International Conference on Secure Cyber Computing
  and Communication (ICSCCC)}, pages 35--38, 2018.

\bibitem{Support1}
Xin Wu, Yuchen Gao, and Dian Jiao.
\newblock Multi-label classification based on random forest algorithm for
  non-intrusive load monitoring system.
\newblock {\em Processes}, 7(6), 2019.

\bibitem{marlen}
Cecilia Panigutti, Riccardo Guidotti, Anna Monreale, and Dino Pedreschi.
\newblock Explaining multi-label black-box classifiers for health applications.
\newblock In {\em International Workshop on Health Intelligence}, pages
  97--110. Springer, 2019.

\bibitem{Tabia}
Karim Tabia.
\newblock Towards explainable multi-label classification.
\newblock In {\em 2019 IEEE 31st International Conference on Tools with
  Artificial Intelligence (ICTAI)}, pages 1088--1095, 2019.

\bibitem{lime}
Marco~Tulio Ribeiro, Sameer Singh, and Carlos Guestrin.
\newblock Why should i trust you?: Explaining the predictions of any
  classifier.
\newblock In {\em Proceedings of the 22nd ACM SIGKDD international conference
  on knowledge discovery and data mining}, pages 1135--1144. ACM, 2016.

\bibitem{anchors}
Marco~Tulio Ribeiro, Sameer Singh, and Carlos Guestrin.
\newblock {Anchors: High-Precision Model-Agnostic Explanations}.
\newblock In {\em Thirty-Second AAAI Conference on Artificial Intelligence},
  2018.

\bibitem{riccardo_guidotti1}
Riccardo Guidotti, Anna Monreale, Fosca Giannotti, Dino Pedreschi, Salvatore
  Ruggieri, and Franco Turini.
\newblock Factual and counterfactual explanations for black box decision
  making.
\newblock {\em IEEE Intelligent Systems}, 34(6):14--23, 2019.

\bibitem{trepan}
Mark Craven and Jude Shavlik.
\newblock Extracting tree-structured representations of trained networks.
\newblock In D.~Touretzky, M.C. Mozer, and M.~Hasselmo, editors, {\em Advances
  in Neural Information Processing Systems}, volume~8. MIT Press, 1995.

\bibitem{rulefit}
Jerome~H Friedman and Bogdan~E Popescu.
\newblock Predictive learning via rule ensembles.
\newblock {\em The Annals of Applied Statistics}, 2(3):916--954, 2008.

\bibitem{moore}
Alexander Moore, Vanessa Murdock, Yaxiong Cai, and Kristine Jones.
\newblock Transparent tree ensembles.
\newblock In {\em The 41st International ACM SIGIR Conference on Research \&
  Development in Information Retrieval}, SIGIR '18, page 1241–1244, New York,
  NY, USA, 2018. Association for Computing Machinery.

\bibitem{iForest}
Xun Zhao, Yanhong Wu, Dik~Lun Lee, and Weiwei Cui.
\newblock iforest: Interpreting random forests via visual analytics.
\newblock {\em IEEE Transactions on Visualization and Computer Graphics},
  25(1):407--416, 2019.

\bibitem{ExMatrix}
M{\'{a}}rio~Popolin Neto and Fernando~V. Paulovich.
\newblock Explainable matrix - visualization for global and local
  interpretability of random forest classification ensembles.
\newblock {\em {IEEE} Trans. Vis. Comput. Graph.}, 27(2):1427--1437, 2021.

\bibitem{foodtruck}
Adriano Rivolli, Larissa~C Parker, and Andre~CPLF de~Carvalho.
\newblock Food truck recommendation using multi-label classification.
\newblock In {\em EPIA Conference on Artificial Intelligence}, pages 585--596.
  Springer, 2017.

\bibitem{waterquality}
Hendrik Blockeel, Sa{\v{s}}o D{\v{z}}eroski, and Jasna Grbovi{\'c}.
\newblock Simultaneous prediction of multiple chemical parameters of river
  water quality with tilde.
\newblock In {\em European Conference on Principles of Data Mining and
  Knowledge Discovery}, pages 32--40. Springer, 1999.

\bibitem{uci}
Dheeru Dua and Casey Graff.
\newblock {UCI} machine learning repository, 2017.

\bibitem{bigrules}
Alex~A. Freitas.
\newblock Comprehensible classification models: A position paper.
\newblock {\em SIGKDD Explor. Newsl.}, 15(1):1–10, March 2014.

\end{thebibliography}

\end{document}